\documentclass[letterpaper, 10 pt, conference]{ieeeconf}  
\usepackage{graphicx}
\usepackage{enumerate}
\usepackage{balance}
\usepackage{subfig}
\usepackage{tabularx}
\usepackage{multirow}

\IEEEoverridecommandlockouts                              

\title{\LARGE \bf
Fusing Saliency Maps with Region Proposals for Unsupervised Object Localization
}

\author{Hakan Karao{\u{g}}uz$^{1}$ and Patric Jensfelt$^{1}$
\thanks{$^{1}$ The authors are all with the Centre for Autonomous Systems at
KTH Royal Institute of Technology, Stockholm, SE-100 44, Sweden {\tt\small hkarao@kth.se}}%
}

\begin{document}

\maketitle
\thispagestyle{empty}
\pagestyle{empty}

\begin{abstract}

In this paper we address the problem of unsupervised localization of objects in single images. Compared to previous state-of-the-art method our method is fully unsupervised in the sense that there is no prior instance level or category level information about the image. Furthermore, we treat each image individually and do not rely on any neighboring image similarity. We employ deep-learning based generation of saliency maps and region proposals to tackle this problem. First salient regions in the image are determined using an encoder/decoder architecture. The resulting saliency map is matched with region proposals from a class agnostic region proposal network to roughly localize the candidate object regions. These regions are further refined based on the overlap and similarity ratios. Our experimental evaluations on a benchmark dataset show that the method gets close to current state-of-the-art methods in terms of localization accuracy even though these make use of multiple frames. Furthermore, we created a more challenging and realistic dataset with multiple object categories and varying viewpoint and illumination conditions for evaluating the method's performance in real world scenarios.

\end{abstract}

\section{INTRODUCTION}

Autonomous systems encounter a variety of objects in real-world scenarios. Reasoning about and interacting with the environment often involves localizing or recognizing these objects.

Object  recognition \cite{lowe1999object} and localization \cite{sullivan1999object} have been extensively studied in the literature. 
The traditional pipeline for these tasks is to collect data, train a supervised classifier and evaluate the system on a benchmark dataset. With the recent advancements in deep-learning based classifiers such as Convolutional Neural Network (CNN), significant improvements in recognition/detection accuracy have been achieved both for benchmark datasets and real-time systems. Moreover, large datasets have emerged \cite{imagenet,microsoft_coco} to train and evaluate these data-hungry networks.

\begin{figure}[h!tb]
 \centering
 \includegraphics[width=0.98\columnwidth]{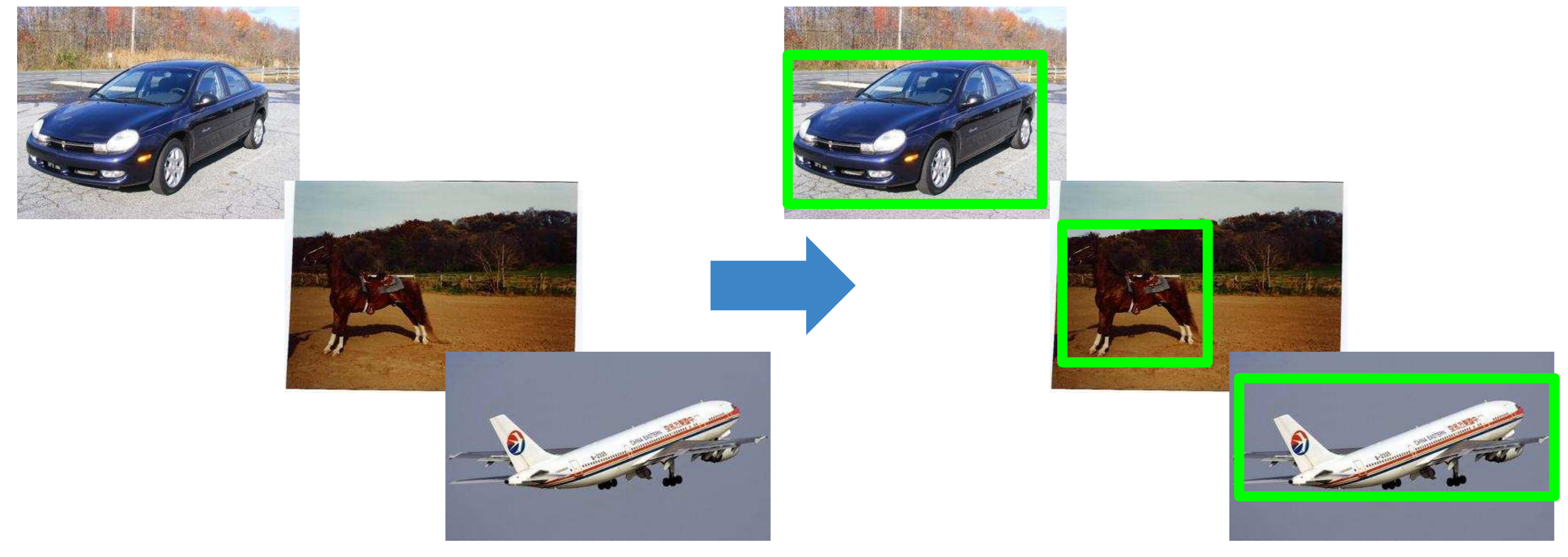}
 \caption{The problem of object localization. Left: Input of unannotated single images that contain objects of unknown categories. Right: The desired output which is bounding-box localization (show in green) of these objects.}
 \label{fig:goal}
 \end{figure}

However, even these large datasets are limited to a few hundreds of object categories when it comes to object detection, localization and segmentation tasks. This is because labeling of the data is costly and time consuming. Lately, methods such as Generative-Adversarial-Network (GAN) are employed to generate and enhance additional data from small datasets \cite{Salimans2016}. However, these methods are still far from generating a large generic and realistic dataset from a small set of labeled data. 

As a result, current state-of-the-art supervised object recognition/detection methods are limited to the categories present in the training data. Moreover these pretrained classifiers often work only when the objects are seen from some common viewpoints. 
Fine-tuning is generally applied to overcome these issues. However, as mentioned before data collection and labeling is costly and time consuming task and the classifier needs to be retrained for every new object category.

In this work we study the object localization problem in a completely unsupervised setting. Our aim is a system that receives an unannotated image and outputs bounding-boxes that most likely contain an object. Our motivation is that robots deployed in challenging environments such as factories, mines, construction sites etc, can encounter custom made, rarely seen objects of any size and shape such as specialized machineries, custom made bolts and tools. Moreover, the environmental conditions can vary a lot. A mine is usually much darker than a regular office floor or the illumination in a manufacturing plant may vary rapidly during a welding process etc. Therefore, it is very hard to address all these issues beforehand and collect enough data to train a supervised classifier. An unsupervised approach on the other hand has the potential to locate objects in a scene without any prior, object specific,  information.

Our problem is similar to some of the previously studied problems in the literature such as weakly supervised object discovery \cite{cinbis2014multi}. There, the images are labeled as positive or as negative samples for each object class without any bounding box annotations. There are also connections to the colocalization problem \cite{deselaers2010}, where the aim is to localize a single dominant object class in a set of valid images and some noisy images without an object of the same class. In our setting, we assume there is at least one object of arbitrary category present in each image and the object needs not to be dominating the scene.  

Our main contribution is a method that combines pixel-wise saliency maps that are obtained from a deep semantic segmentation network and region proposals generated with a class agnostic region proposal network (RPN). This allows us to detect potential objects efficiently from single images without any additional information. Moreover, we contribute a dataset that is collected from a robot platform. The images are collected with varying illumination/viewpoint conditions to allow for a more in-depth analysis of the real-world robotics performance of the method. We also evaluate our method on a benchmark dataset and show that we perform on par with current state-of-the-art methods despite using only a single frame only and not using similarities from neighboring images.

The organization of this paper is as follows. The related work on unsupervised object discovery methods is introduced in Section~\ref{sec:rellit}. In Section~\ref{sec:background} background about common unsupervised object localization methods is given and in Section~\ref{sec:method}, our approach is explained in detail. Our contributed dataset is presented in Section~\ref{sec:handtooldataset}. The experimental results are given in Section~\ref{sec:exp} and the paper concludes with a brief summary and conclusions and suggestions for future work in Section~\ref{sec:conclusion}.

\begin{figure}[!ht]
      \centering
      \includegraphics[width=0.9\columnwidth]{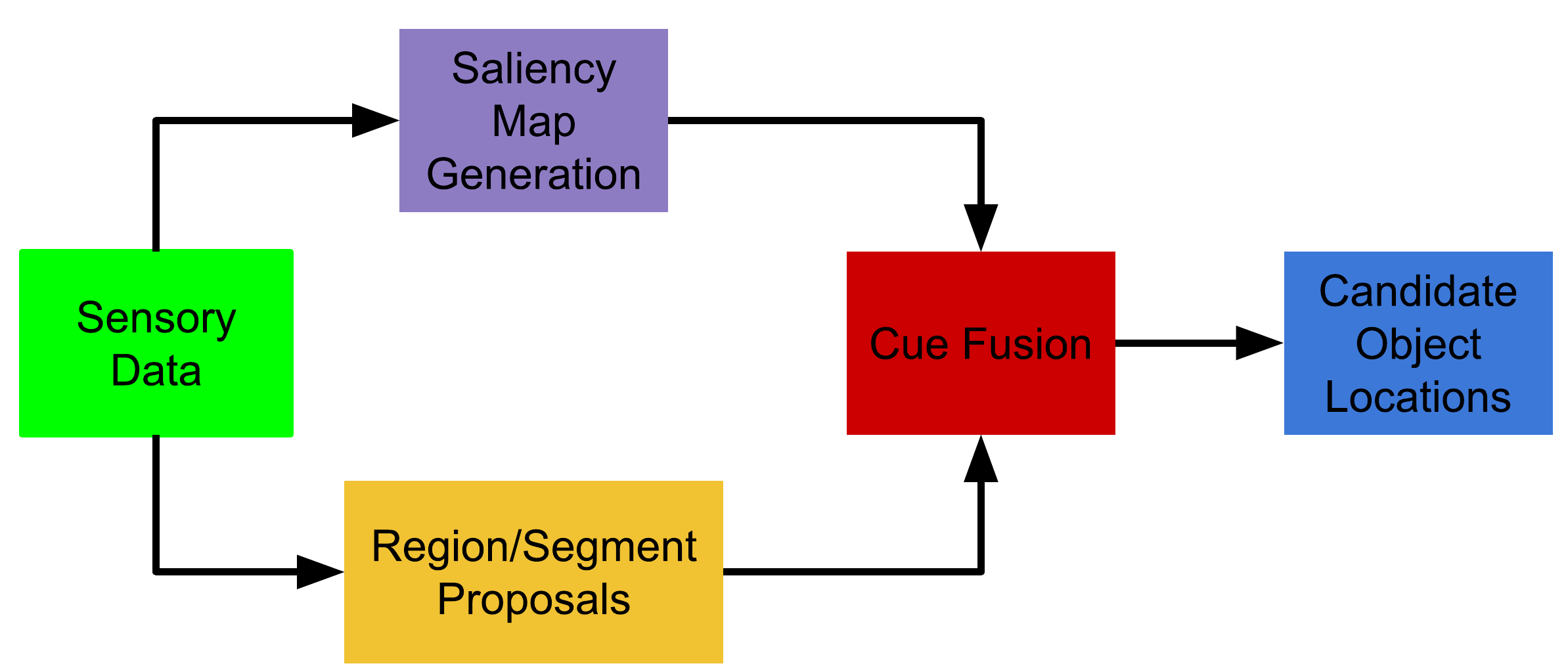}
      \caption{Overview of the common unsupervised object localization pipeline. }
      \label{fig:method}
 \end{figure}

\begin{figure*}[!htb]
      \centering
      \includegraphics[width=0.8\textwidth]{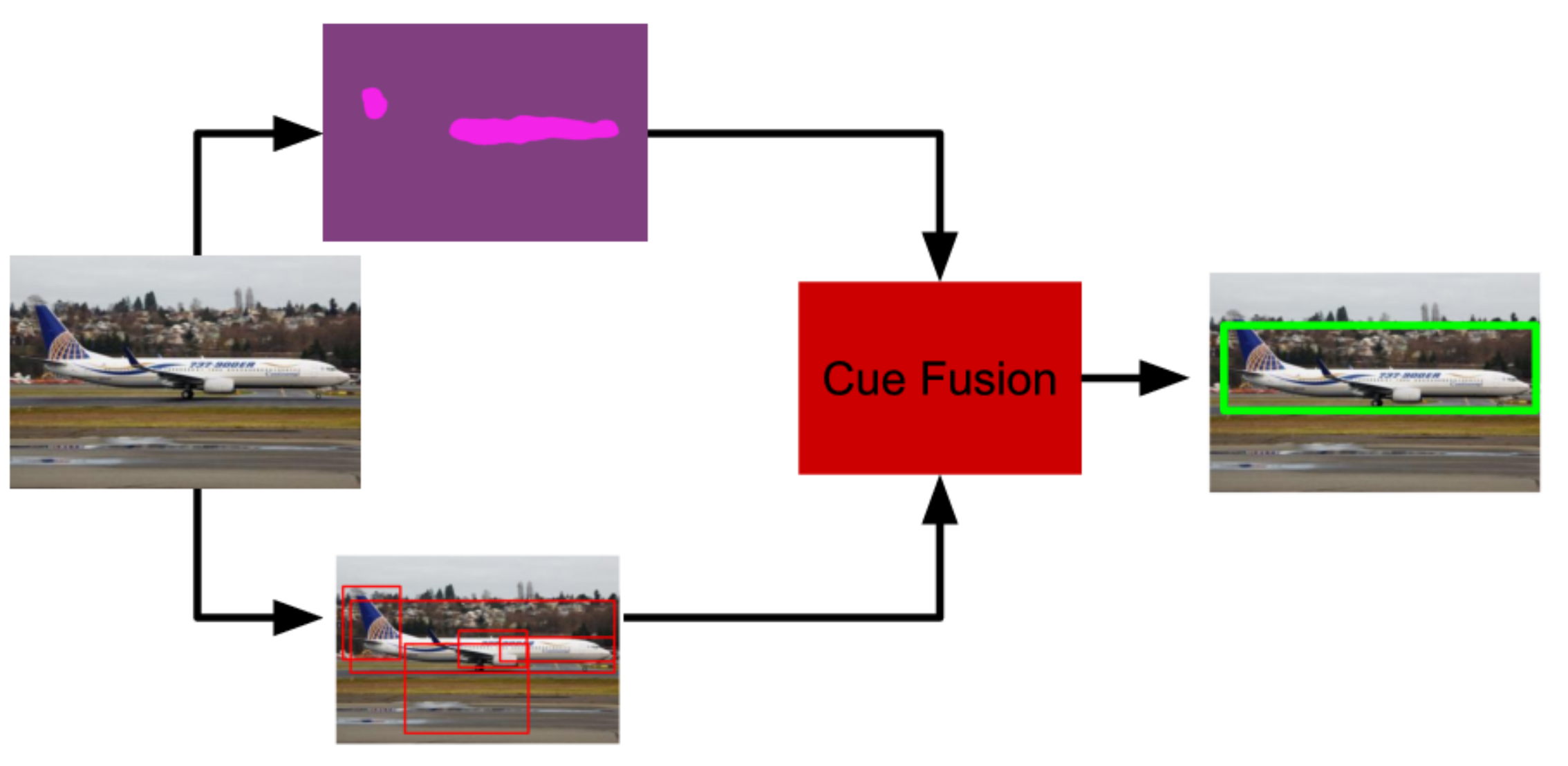}
      \caption{The example outputs of the individual modules at each step. }
      \label{fig:impl}
 \end{figure*}

\section{Related Work}
\label{sec:rellit}

Unsupervised object discovery, localization and segmentation are highly interleaved topics since the end-goal for all of these tasks is to discover and extract objects in a scene without any supervision.  The difference between the approaches comes from the context where the solution is applied. Some work \cite{bjorkman2010active,kootstra2011fast,potapova2014attention,ma2015simultaneous,Abbeloos20173DOD} focus on real robot scenarios with the end goal of locally segmenting objects. While others \cite{russell2006using,felzenszwalb2004efficient,Rubinstein2013, Ghafarianzadeh2014,Cho2015} focus on image collections or video sequences and segment the images globally.

Unsupervised object segmentation approaches use various cues such as single images \cite{felzenszwalb2004efficient}, stereo vision  \cite{bjorkman2010active, kootstra2011fast, Kootstra2010,mishra2009active}, RGB-D \cite{potapova2014attention,Abbeloos20173DOD}, video \cite{grundmann2010,Ghafarianzadeh2014} and even robot perturbation \cite{ma2015simultaneous}.  Such approaches can be divided into two categories \cite{potapova2014attention}. 

The first category is discriminative approaches \cite{felzenszwalb2004efficient,Ghafarianzadeh2014} where an image is treated as a whole and the goal is to globally segment the image into regions. Although these approaches are not primarily concerned with discovering individual objects in an image, they serve as a basis for this task. Felzsenswalb et al. \cite{felzenszwalb2004efficient} create undirected graphs for several image features and compare them to create boundaries between different segments. Grundmann et al. \cite{grundmann2010} improve the graph based segmentation for video sequences using a hierarchical approach. Finally, Ghafarianzadeh et al. \cite{Ghafarianzadeh2014} adds spatio-temporal information to further improve graph based segmentation accuracies for video sequences. This method is employed by Ma et al. \cite{ma2015simultaneous} to get initial object candidates in a scene. Thereafter, a candidate segment is randomly chosen and is attempted to be manipulated within a predefined trajectory in order to validate and model it.    

The second category is agglomerative approaches \cite{bjorkman2010active,kootstra2010fixation,Kootstra2010,potapova2014attention,mishra2009active,Horbert2015,garcia2015saliency,Rubinstein2013}. In these approaches, initial segment locations are determined either by using some image features or saliency maps. Afterwards, individual regions are determined by means of pixel-similarity and region growing. The common pipeline for these methods is shown in Fig.~\ref{fig:method}. Bjorkman et al. \cite{bjorkman2010active}, uses an active stereo vision system to detect and segment unknown object that is centered in the view and assumed to stand on a planar surface. They use color and disparity information to generate a rough segmentation of the image to initialize probabilistic models for the background, foreground (object) and planar surface classes. Then they employ the Expectation-Maximization (EM) algorithm to iteratively improve the models of these classes.  Kootstra et al. \cite{kootstra2011fast} use some of the Gestalt principles to generate a saliency map and fixation points in an image. 
Using these fixation points, they employ super-pixel segmentation and clustering to segment the image into foreground and background segments. The generated segments are analyzed further to measure their goodness with a pretrained neural network.
A similar system that uses fixation points as seeds for segmentation is proposed by Mishra et al. in \cite{mishra2009active}. In that work the edge map of an image is used to segment objects that are assumed to be marked by fixation points. The initial edge map is created using color and texture features and it is further refined with motion cues.

In \cite{potapova2014attention}, RGB-D data is employed to detect tabletop objects in a scene. They employ 2.5D symmetry features for generating fixations and segmenting the scene using surface normals and patch similarity. Garcia et al. \cite{garcia2015saliency} uses color and depth cues independently to find the object candidates for both cues and perform late-fusion to refine the results of separate cues.
Abbeloos et al. \cite{Abbeloos20173DOD} use SIFT correspondences and geometric verification to discover the instances of an unknown object in cluttered scenes. The disadvantage of this method is that it requires at least two instances of an object to be present in the scene. 

Russel et al. \cite{russell2006using} applied multi-segmentation and histogram of visual words together with a probabilistic Latent Dirichlet Allocation model to discover and segment objects in an image dataset. Rubinstein et al. \cite{Rubinstein2013} employed saliency along with dense correspondences between images to discover and segment common objects. 

For the problem of unsupervised object localization, there are important works \cite{Tang2014,joulin2014efficient,Cho2015} that perform bounding-box localization on image collections.  Tang et al. \cite{Tang2014} perform colocalization on images by creating convex image and bounding-box models seperately and perform convex optimization to find the optimal bounding-box that contains the object in an image as well as the noisy images that do not contain the object. Joulin et al. \cite{joulin2014efficient} extend these models for colocalization in videos with temporal constraints.
Cho et al. \cite{Cho2015} use an off-the-shelf object proposal approach combined with probabilistic Hough matching in an iterative way to localize objects. With their method, they achieve state-of-the art-results for object localization on benchmark datasets.

Our work is along the lines with \cite{Tang2014,joulin2014efficient,Cho2015}, with a pipeline like in Fig.~\ref{fig:method}. However the primary difference of our method is that our method can do one-shot bounding-box localization in a single image whereas the above methods need a set of images to propose reasonable localization. Moreover, different from the previous work we employ a pixel-wise deep semantic segmentation network for saliency map generation and employ a class agnostic region proposal network for generating bounding-box proposals. We have evaluated our approach both on the benchmark dataset used in \cite{Rubinstein2013,Cho2015} and on a dataset collected from a real robot with explicit changes in the environment such as illumination and viewpoint with challenging objects to get a more in-depth analysis about the system's performance.


\section{Background}
\label{sec:background}
In this section, we briefly outline the individual components of the common unsupervised object localization pipeline which is shown in Fig.~\ref{fig:method}, as a way to lay the ground for explaining our method in the next Section.

\subsection{Saliency Map Generation}
Visual saliency is used to locate and process only the attentive regions of an image \cite{Borji15}. Some of the approaches try to locate salient objects or contexts in the image while others estimate where a human eye would fixate as in indirect way to achieve the same. Formally, given an image $I \in \mathcal{I}$, the output of the saliency map is a grayscale image where the intensity values indicate how salient a pixel is. When the grayscale image is normalized it becomes a probability map $p_s(I)$, where for each pixel $(x,y)$ a saliency measure is given as a value $p_s(x,y)$ such that $0 \leq p_s(x,y) \leq 1$. This output is particularly useful for unsupervised object localization task.

\subsection{Region Proposals}
Region proposal methods generate bounding-box proposals in an image, ideally centered around potential objects or salient areas. Previously, sliding window approaches were used to generate proposals which was slow and inaccurate. Later, \cite{alexe2012objectness} proposed an objectness method to generate proposals based on image features. Recently, deep-learning based proposal methods are developed for generating accurate bounding-box proposals \cite{renNIPS15fasterrcnn}. Formally, given an image $I \in \mathcal{I}$, the output of the region proposal function are bounding-boxes $\sum_{n=1}^{N}$ $rp_{n_I}$, where $N$ is generally a predefined parameter in the order of thousands and $rp_{n_I}=(x_1,y_1,x_2,y_2)$  is a rectangular region with top left anchor $(x_1,y_1)$ and bottom right anchor $(x_2,y_2)$. 

\subsection{Cue Fusion}
\label{sec:cuefusion}
Developing a method for fusing the saliency and region proposal cues is not trivial and generally application specific. In Cho et al. \cite{Cho2015}, part-based region matching is fused with contrast standout measure to refine region proposals. Tang et al. \cite{Tang2014}, refine proposals gathered from objectness approach with an off-the-shelf saliency map. Formally, for each image $I$, the task is to output candidate object boxes $\sum_{k=1}^{K} o_{k_I}$ refined from region proposal set $o_{k_I} \in rp_I$. The standard approach is getting the most salient areas from the saliency map first and then using the centroids of these salient regions as fixation points. With this information, regions that do not contain any fixation could be eliminated. The further refinement on the remaining boxes can be done by checking their overlaps, applying non-maximum suppression, etc.

\begin{table*}[htb]
\begin{center}
 \caption{Example images from KTH Handtool dataset} \label{tab:dataset}
  \begin{tabularx}{0.8\textwidth}{XXXX} 
     \hline  \multirow{2}{*}{Camera Type}  & \multicolumn{3}{c}{Illumination}  \\ \cline{2-4}
      & \hspace{5mm} Artificial & \hspace{5mm} Natural & \hspace{5mm} Directional  \\ 
      \hline  \vspace{6mm} $Camera_1$ & \vspace{0.5mm} \parbox[c]{1em}{\includegraphics[width=1in]{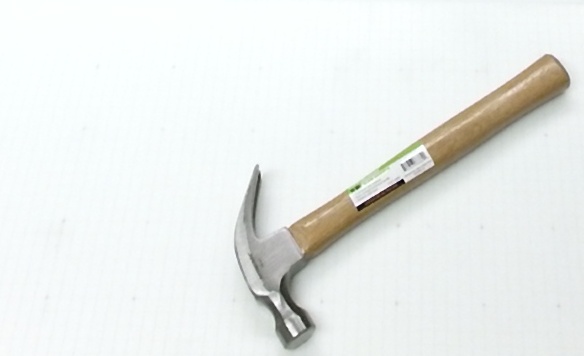}} & 
      \vspace{0.5mm} \parbox[c]{1em}{\includegraphics[width=1in]{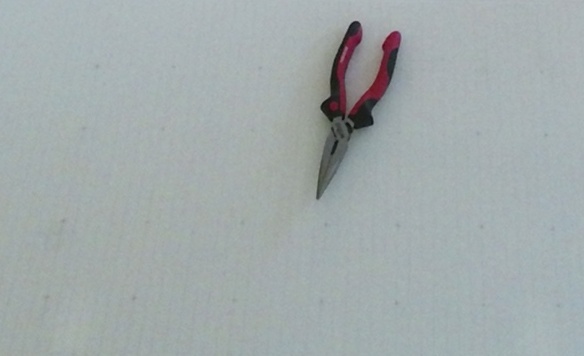}} & 
      \vspace{0.5mm} \parbox[c]{1em}{\includegraphics[width=1in]{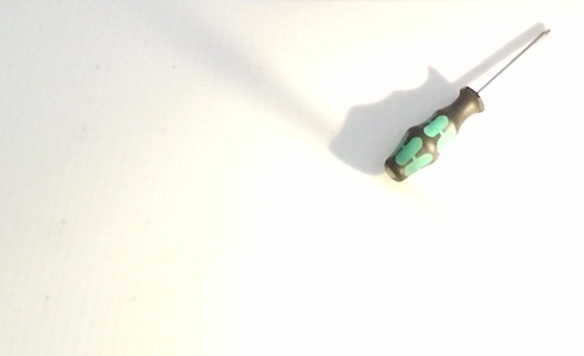}} \\
        \vspace{6mm} $Camera_2$ & \vspace{0.5mm} \parbox[c]{1em}{\includegraphics[width=1in]{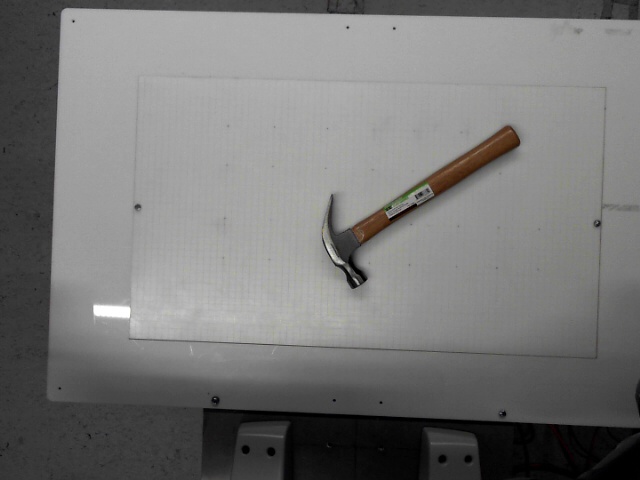}}  & \vspace{0.5mm} \parbox[c]{1em}{\includegraphics[width=1in]{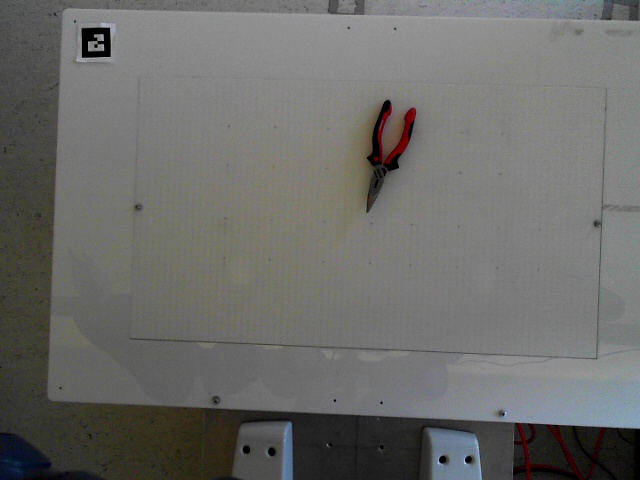}} & \vspace{0.5mm} \parbox[c]{1em}{\includegraphics[width=1in]{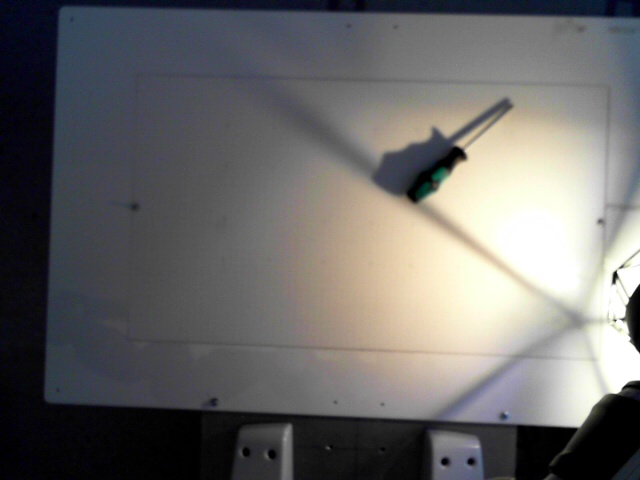}}  \\    
       \hline
  \end{tabularx}
\end{center}
\end{table*}

\section{Method}
\label{sec:method}
Given an RGB image $I \in \cal{I}$, the goal is to locate an unknown number of objects $o_i \in O_{I}$ and mark these with bounding-boxes. Similar to previous work, we have used the common pipeline shown in Fig.~\ref{fig:method}. However, different from the previous work we have employed two seperate deep neural networks for saliency map generation and region proposals. Compared to previous work our system is fully unsupervised and requires only single images.

In order to generate the saliency maps, we employ a Dilated Residual Network (DRN) architecture which is originally designed for image recognition and pixel-wise semantic segmentation \cite{Yu2017}. The reason for choosing this architecture is that by using dilations, the network has higher resolution feature maps and better performance compared to standard ResNet architectures for semantic segmentation tasks. For training the network we have used the saliency prediction data from LSUN 2017 challenge \cite{Yu2015}. In order to get a pixel-wise saliency map output from the network, we require finite number of classes. The traditional saliency maps are 8-bit grayscale images in which the level of intensity (0-255) measures the saliency \cite{Borji15}. It is not practical to treat each level of intensity as a seperate class. Instead we have binarized the graysale saliency maps in the dataset using a threshold of $t_{p_s} = 127$. If the intensity value of a pixel is above the threshold, it is marked as salient, otherwise it is marked as non-salient. Thus, our resulting saliency maps have binary output as seen in Fig.\ref{fig:impl} (top image) where the dark purple colored pixels are non-salient while solid pink pixels are salient.

For generating the region proposals, we employ a class agnostic region proposal network (RPN) mentioned in \cite{dai16rfcn}. This RPN which is originally proposed in \cite{renNIPS15fasterrcnn} is a shallow network that is designed to generate region proposals by using the convolutional feature maps. Smooth $L_1$ regression loss is used during the training of the network. The advantages of this RPNs are that, i) it is translation invariant which is very important since our aim is to be able to localize an object even if it translates in the image and ii) it embeds the multi-scale bounding box regression through anchors which is desirable for localizing different sized objects. Example bounding-box proposals obtained from this RPN is shown in Fig.~\ref{fig:impl} (bottom image).

For the cue fusion part, we use the standard approach explained in Sec.~\ref{sec:cuefusion}. We start by refining the saliency map by employing an area threshold $t_a$ to remove the salt-pepper like noisy salient regions. For the remaining salient regions, we compute the centroids and store them as fixation points. The region proposals which do not contain any fixation points are discarded. Then we perform non-maximum suppression with threshold $t_{nms}$ to the remaining region proposals for pruning highly overlapping ones. If there are any low overlapping boxes, we perform color histogram comparison based on threshold $t_{hist}$ for deciding whether to join them or not. The final output of this method is the candidate object regions.

Fig.~\ref{fig:impl} shows outputs of our modules for the common unsupervised object localization pipeline.

\section{KTH Handtool dataset}
\label{sec:handtooldataset}
The KTH Handtool dataset \cite{kth_handtool} is collected for evaluating the object recognition/detection performance of computer vision methods in varying viewpoint, illumination and background settings. It consists of 9 different hand tools from 3 different categories; hammer, plier and screwdriver. The images are collected with a 2-arm stationary robot platform. Two different cameras are used in 3 different illumination conditions. For each hand tool, approximately 40 images with different poses are collected for each camera and illumination setting. Approximately 2200 images are available in the dataset. Table~\ref{tab:dataset} shows example images from the dataset.

\begin{table*}[ht]

\begin{center}
 \caption{Example results from Object Discovery dataset} \label{tab:objdis_respic}
  \begin{tabularx}{0.8\textwidth}{XXXX} 
     \hline  Result  & \multicolumn{3}{c}{Images}  \\ \cline{2-4}   
      \hline  Success & \parbox[c]{1em}{\includegraphics[width=0.17\textwidth]{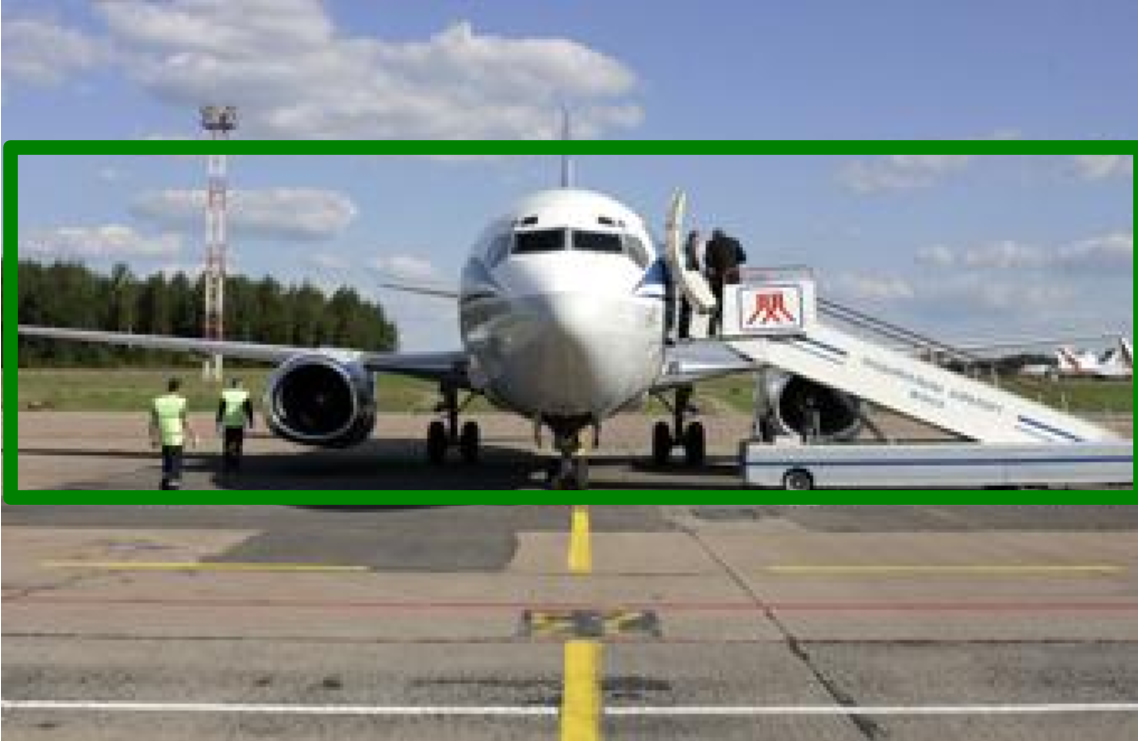}} & \parbox[c]{1em}{\includegraphics[width=0.14\textwidth]{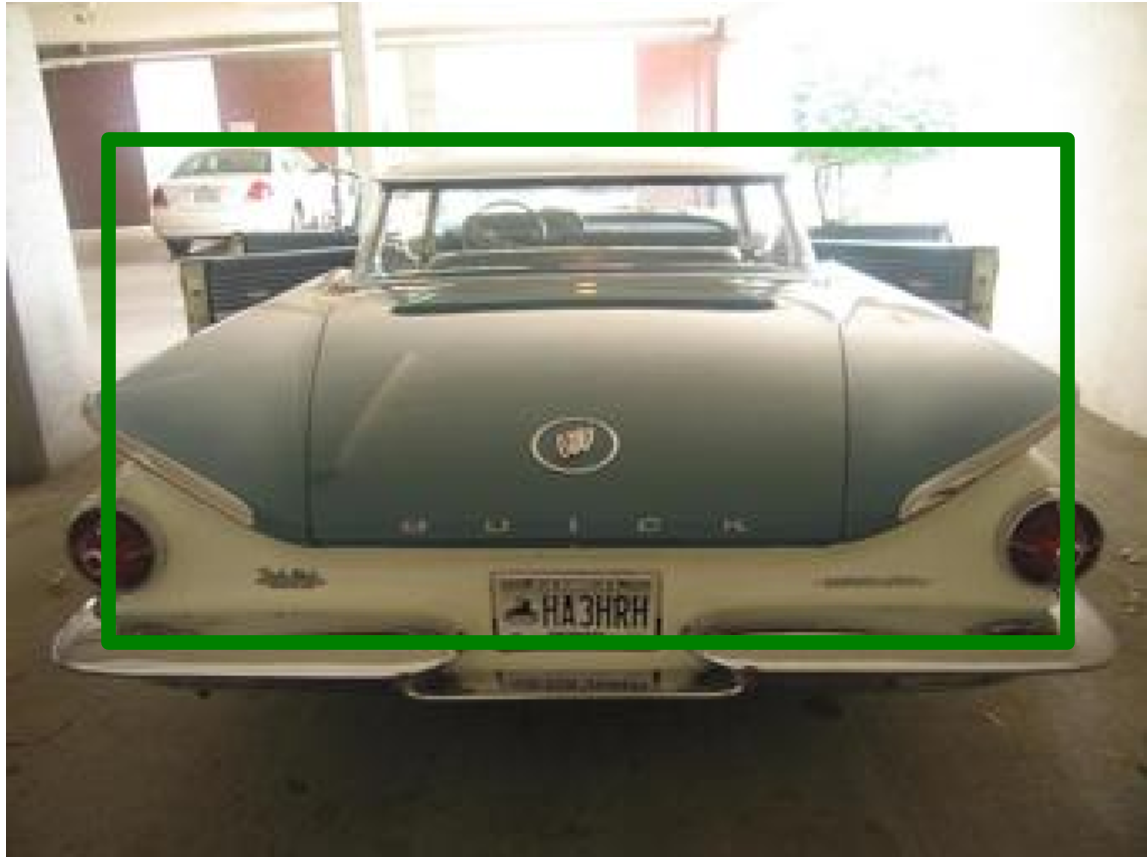}} & \parbox[c]{1em}{\includegraphics[width=0.11\textwidth]{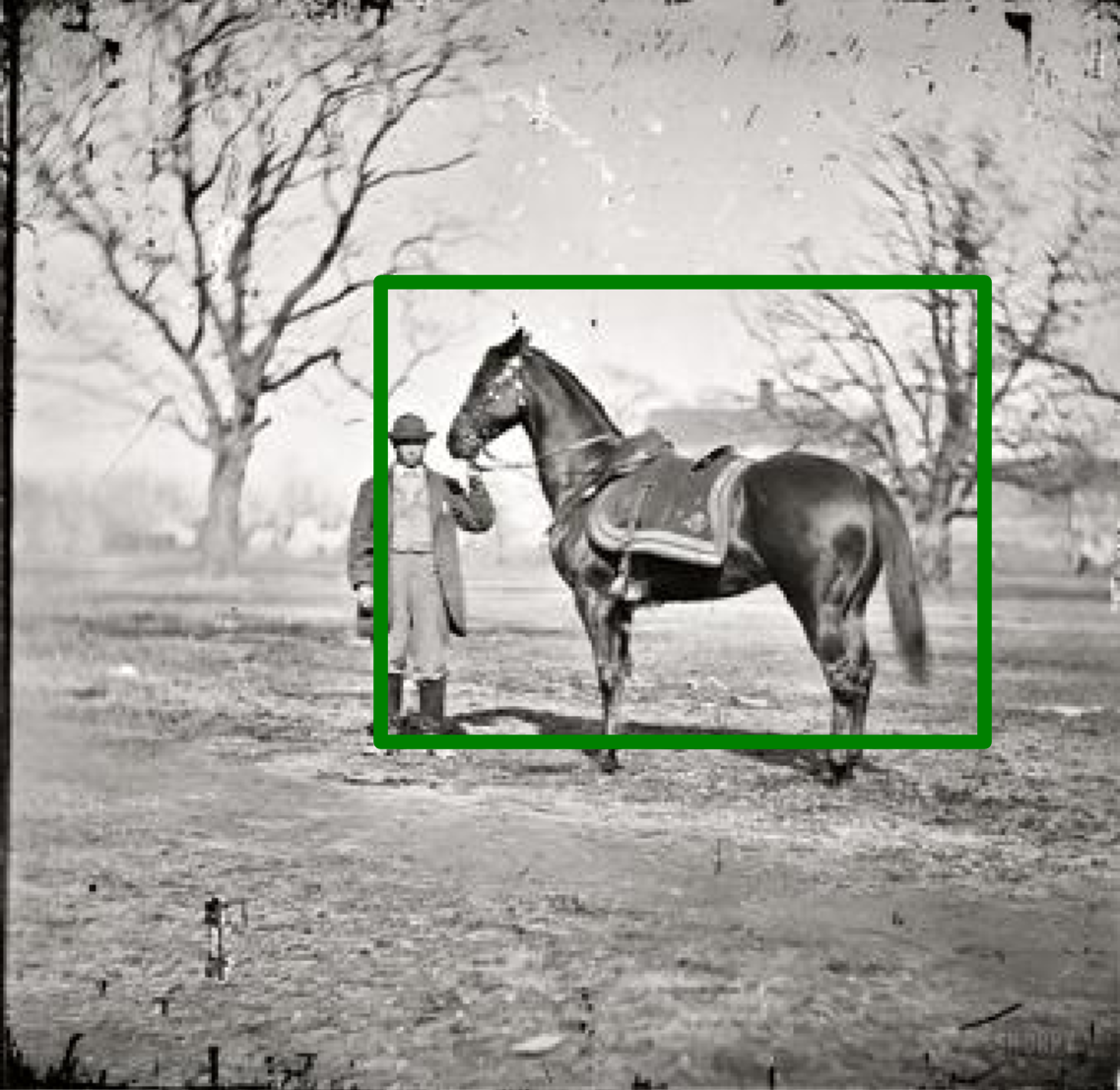}}
       \\ \hline \\
        Fail (Different Object) & & \parbox[c]{1em}{\includegraphics[width=0.17\textwidth]{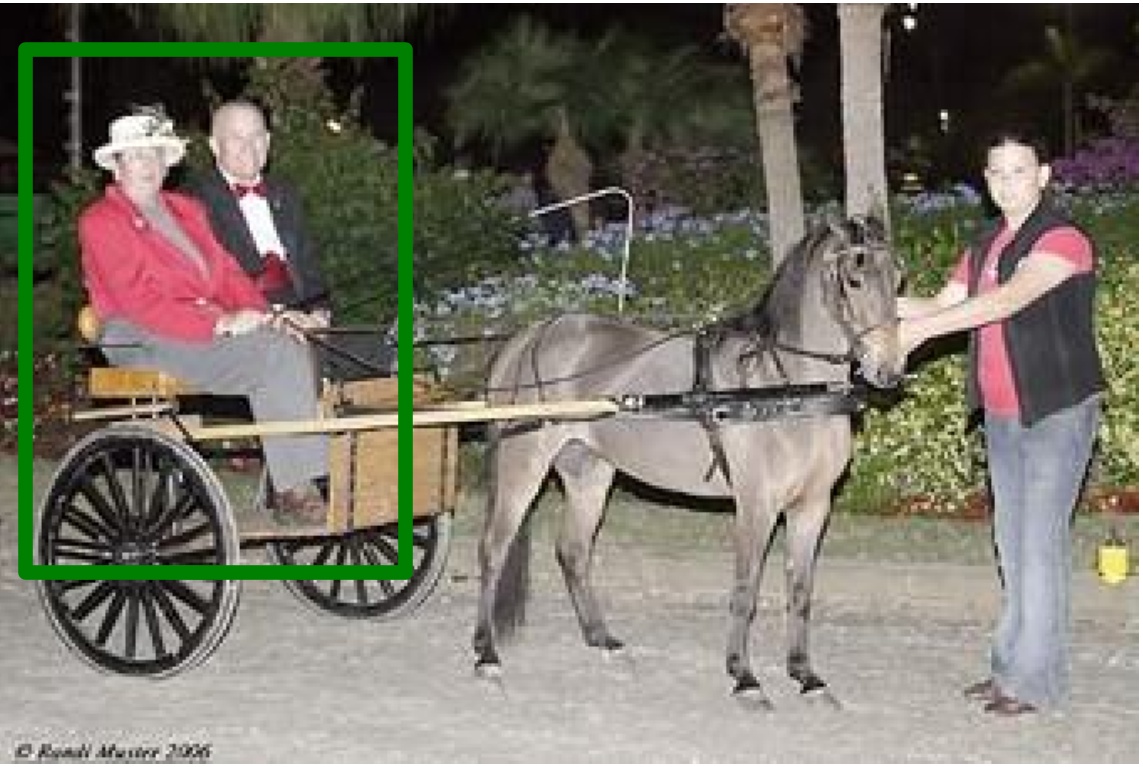}}  &  \\    
       \hline \\
       Fail (Multiple Objects) & \parbox[c]{1em}{\includegraphics[width=0.17\textwidth]{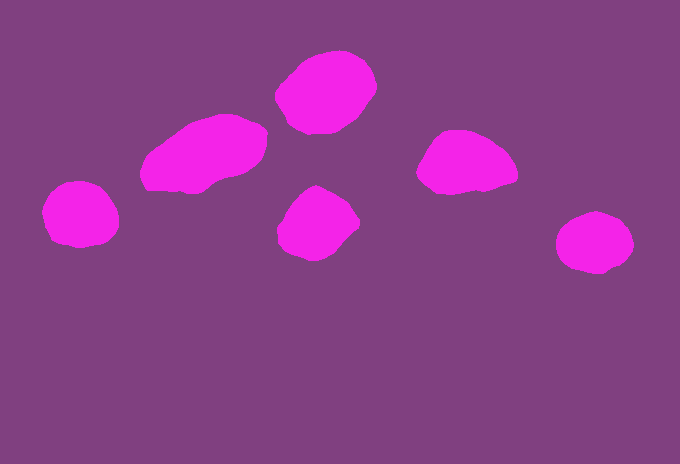}} & \parbox[c]{1em}{\includegraphics[width=0.17\textwidth]{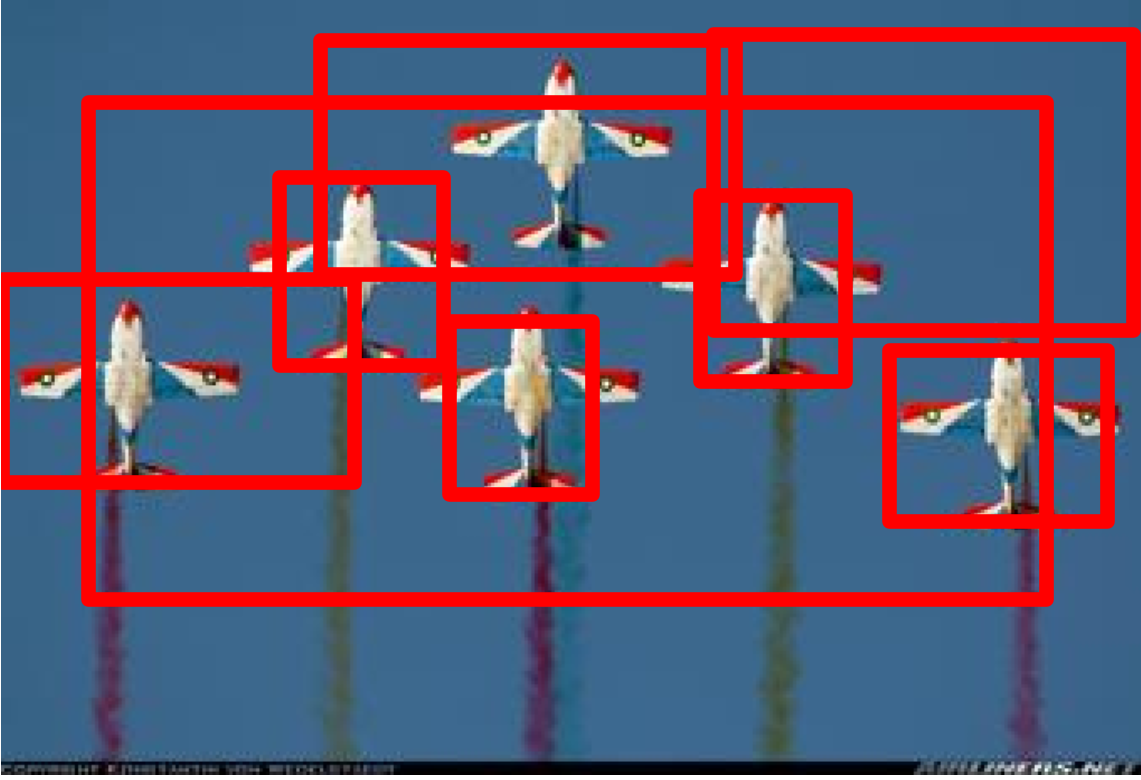}}  & \parbox[c]{1em}{\includegraphics[width=0.17\textwidth]{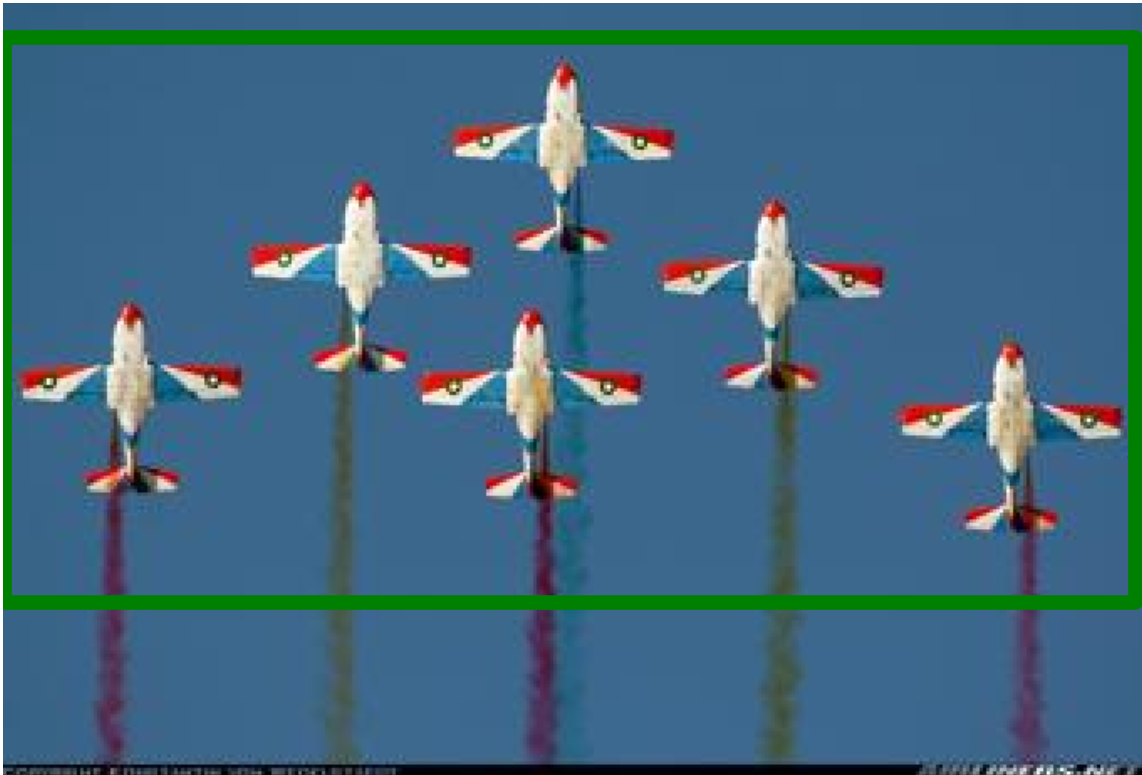}}    \\   \hline
  \end{tabularx}
\end{center}
\end{table*}

\section{Experiments}
\label{sec:exp}
Our experimental evaluation consists of two parts. In the first we compare to the state-of-the-art methods presented in \cite{Rubinstein2013}, \cite{Cho2015} and \cite{Tang2014} on the same benchmark dataset as they use. In the second part we evaluate our method using our Handtool dataset. As evaluation metrics for the performance we use the correct localization (CorLoc) metric which is based on the Jaccard similarity coefficient $J(A,B) = \frac{A \cap B}{A \cup B }$  \cite{Cho2015}. If $J(A,B) > 0.5$ then the object is considered as localized correctly.

\subsection{Object Discovery Dataset}
The first experiment is performed on the dataset from \cite{Rubinstein2013}. This dataset is collected from internet images. To allow for a fair comparison we use the same subset of objects as in \cite{Cho2015}. This subset has 100 images for each of the 3 categories that are airplane, car and horse. Although the dataset is aimed at an unstructured setting most of the images are centered around the one large object of interest which can be seen in Table.~\ref{tab:objdis_respic} (top row). The parameters used in this experiment are given in Table~\ref{tab:objdis_param} and our localization results together with the other approaches are given in Table \ref{tab:objdis_res}. Some of the successful localizations of our method is given in Table~\ref{tab:objdis_respic} (top row).

\begin{table}[h]
\caption{Parameters used for Object Discovery dataset.}
\label{tab:objdis_param}
\begin{center}
\begin{tabular}{|c|c|}
\hline
Parameter & Value\\
\hline
$t_a$ & 300 (pixel$^2$)\\
\hline
$t_{nms}$ & 0.15 \\
\hline
$t_{hist}$ & 1.0 \\
\hline
\end{tabular}
\end{center}
\end{table}

\begin{table}[htb]
\centering
\caption{CorLoc($\%$) results for Object Discovery dataset.}
\label{tab:objdis_res}
\begin{tabular}{|l|c|c|c|c|}
\hline
Methods           & Airplane       & Car            & Horse          & Average        \\ \hline \hline
Rubinstein et al. \cite{Rubinstein2013} & 74.39          & 87.64          & 63.44          & 75.16          \\ \hline
Tang et al. \cite{Tang2014}      & 71.95          & 93.26          & 64.52          & 76.58          \\ \hline
Cho et al.   \cite{Cho2015}     & 82.93 & 94.38 & 75.27 & 84.19 \\ \hline \hline
Ours              & 81.7           & 86.5           & 62.3           & 76.8           \\ \hline
\end{tabular}
\end{table}

%
%

As seen from the table, our approach does not beat the best performing method but is comparable to the other methods in terms of average accuracy. However, all the other approaches require multiple images and/or a few iterations for achieving the final result. As seen in \cite{Cho2015}, their first iteration scores are around $70\%$. Regarding computational time, the systems have been implemented on slightly different hardware. Our approach takes approximately 1.5 seconds per image on a 4 core computer with a 980 GTX GPU. As a comparison, the state-of-the art approach \cite{Cho2015} reports about an hour for 500 images with 10 core computer which makes approximately 7.2 seconds of processing time per image.

If we look at the individual category results, we see that the we obtain rather high scores for airplane and car while our score for horse is lower. The low performance in horse category is probably because of two main reasons. First, other objects such as humans are present and dominant in some of the horse images which shifts the focus of the saliency module. This case can be seen in Table~\ref{tab:objdis_respic} (middle row). As our system performs one-shot object localization on a single image, the output is reasonable in the sense that it finds meaningful objects but in terms of category accuracy it is a failure. Second, in some images the color of the horse and the background is rather similar which makes it hard for the saliency system to distinguish. 

Apart from individual category failures, another general source of failure is when multiple objects are present in the image. An example of this is shown in Table~\ref{tab:objdis_respic} (bottom row). Here, the saliency map and region proposals successfully determine individual planes but the cue fusion algorithm which is biased towards localizing one large object instead of smaller sparse objects combines individual salient regions into one big region. Deciding whether combining sparse salient regions or not is an important aspect that we will investigate further as a future work.

\subsection{KTH Handtool Dataset}

Our approach is also evaluated for each object in our Handtool dataset. The parameters used in this dataset are given in Table~\ref{tab:kthdata_param}. Table~\ref{tab:kthdata_res} shows the obtained results. Because of space constraints, we show only the average category results instead of individual instance results.

\begin{table}[h]
\caption{Parameters used for KTH Handtool dataset.}
\label{tab:kthdata_param}
\begin{center}
\begin{tabular}{|c|c|}
\hline
Parameter & Value\\
\hline
$t_a$ & 300 (pixel$^2$)\\
\hline
$t_{nms}$ & 0.05 \\
\hline
$t_{hist}$ & 1.0 \\
\hline
\end{tabular}
\end{center}
\end{table}

\begin{table}[htb]
\centering
\caption{CorLoc($\%$) results for KTH Handtool dataset.}
\label{tab:kthdata_res}
\begin{tabular}{|c|c|c|c|c|}
\hline
Camera Type    & Illumination & \multicolumn{3}{c|}{Object Type}  \\ \cline{3-5}
    &       & Hammer      & Plier           & Screwdriver  \\ \hline \hline
\multirow{3}{*}{$Camera_1$} & Artificial & 23.1 &	63.7 &	38.3 \\
\cline{2-5}
 & Natural & 37.5 &	58.6 &	31.0 \\
 \cline{2-5}
  & Directional & 23.0 & 59.0 &	27.5 \\
  \hline \hline
\multirow{3}{*}{$Camera_2$} & Artificial & 34.3 &	59.7 &	29.3 \\
\cline{2-5}
& Natural & 14,3 &	47.5 &	15.7 \\
\cline{2-5}
& Directional & 21.5 &	58.9 &	31.0 \\
\cline{2-5} 
\hline
\end{tabular}
\end{table}

From the results, it can be clearly seen that this dataset is more challenging compared to the Object Discovery dataset since the objects of interest are not dominating the scene and the environmental changes are more obvious. Secondly, we see that the most successfully localized category is plier despite its size being smaller than the other two categories. This is probably due to the more textured structure of the tool compared to hammers and screwdrivers. 

If we look at the camera type, we see that the accuracy with $Camera_1$ is generally higher than $Camera_2$. This is due to $Camera_1$ having higher resolution, closer viewpoint and better illumination control. The images from $Camera_2$ are from farther away and are darker. Objects being farther also cause them to be less dominant in the scene.

The variance in the illumination has different effects for each category and camera. For hammer and screwdriver categories, the illumination variance significantly affects the localization accuracy for both cameras. For the plier category, the accuracy seems less affected. For $Camera_1$, screwdrivers and pliers are localized best with artificial illumination conditions while natural light works best for hammer. This could be because of the shiny surface of the hammers that causes the drop in the accuracy with artificial and directional light. 

For $Camera_2$, plier and hammer categories are localized best with artificial light while surprisingly directional light works best for screwdriver with a small margin. The reason for natural light being worst is probably due to the fact that $Camera_2$ images being darker.

\begin{figure}[!ht]
     \centering
     
         \subfloat{\includegraphics[width=0.3\columnwidth]{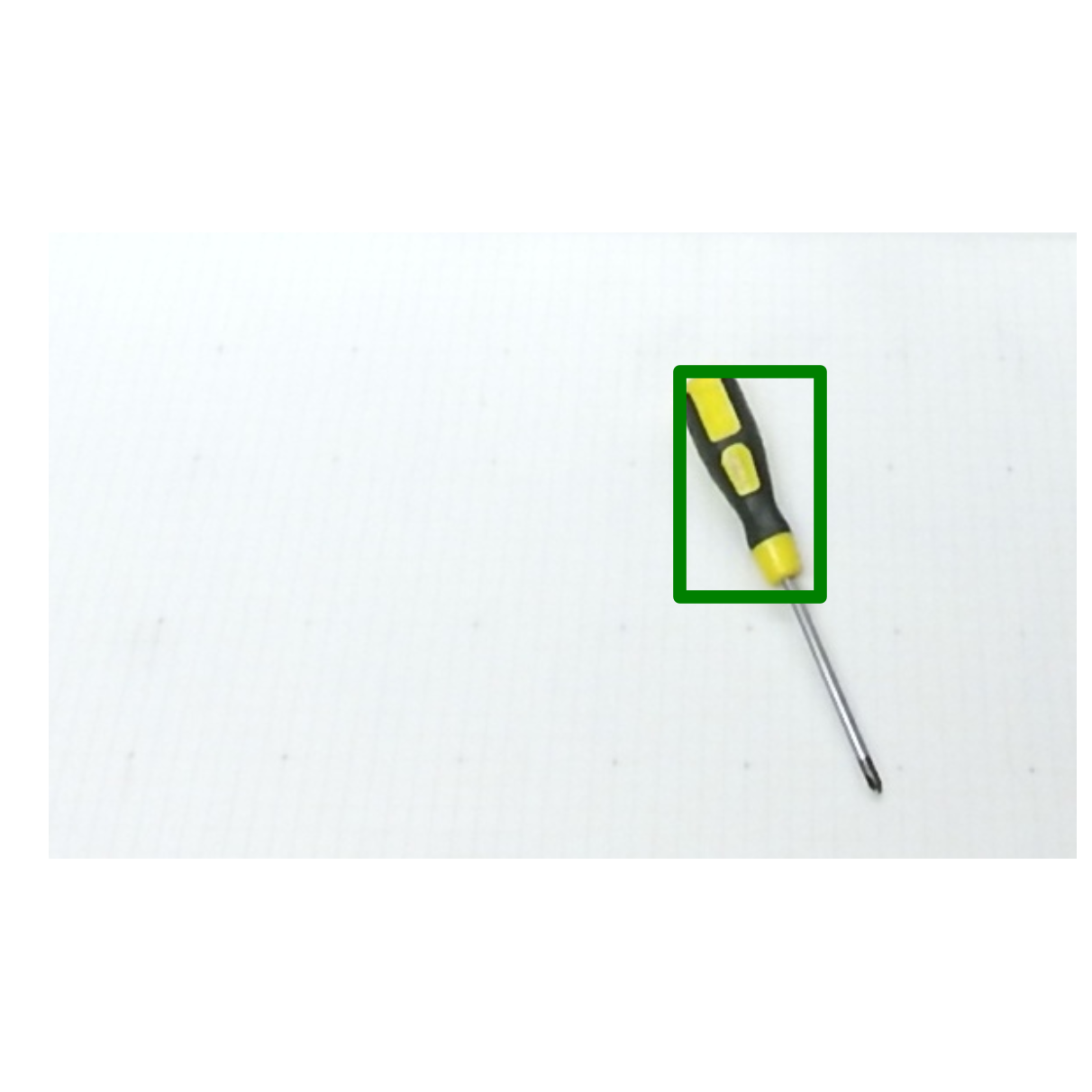}}
         \quad
         \subfloat{\includegraphics[width=0.3\columnwidth]{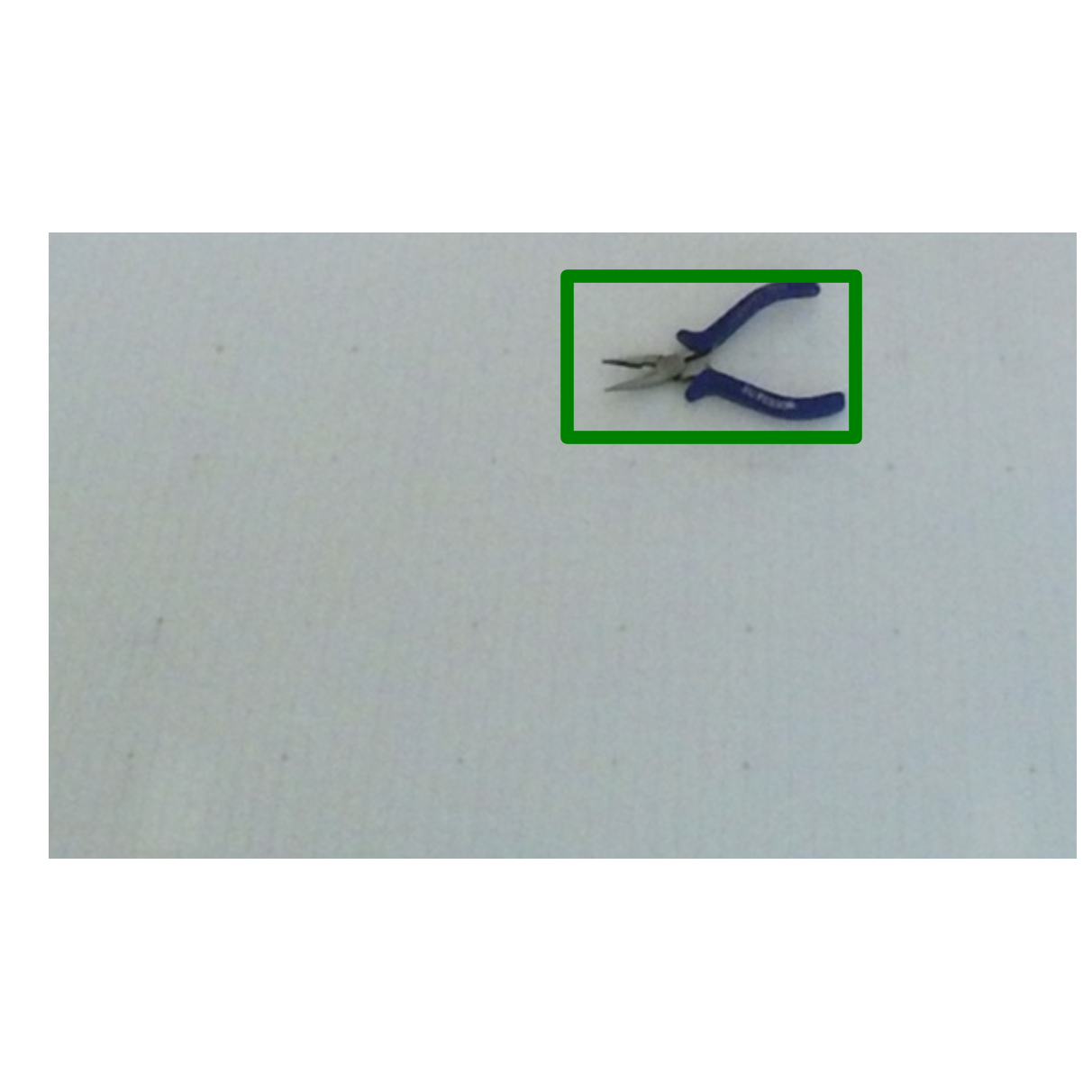}}
         \quad
         \subfloat{\includegraphics[width=0.3\columnwidth]{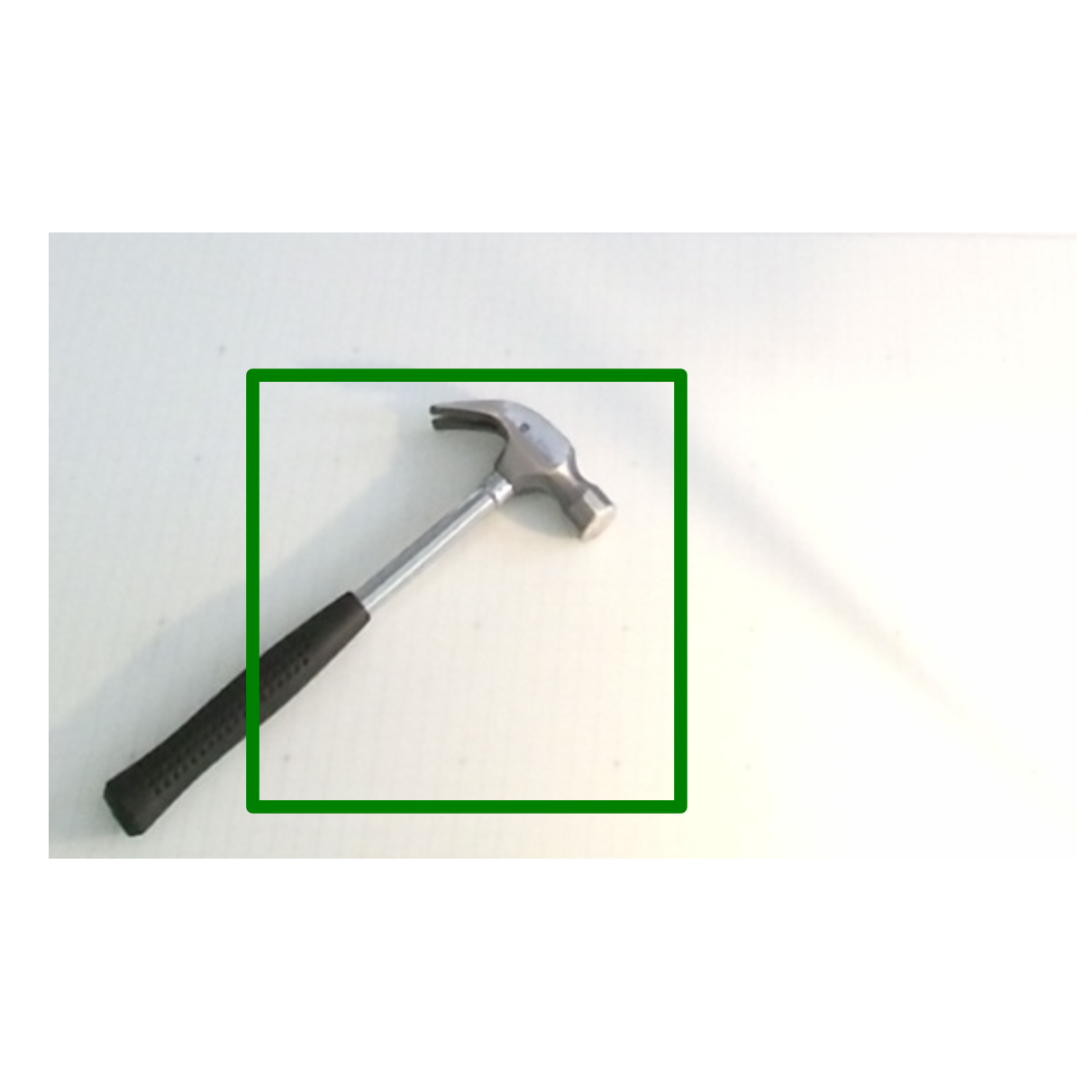}} \\
         
         \subfloat{\includegraphics[width=0.3\columnwidth]{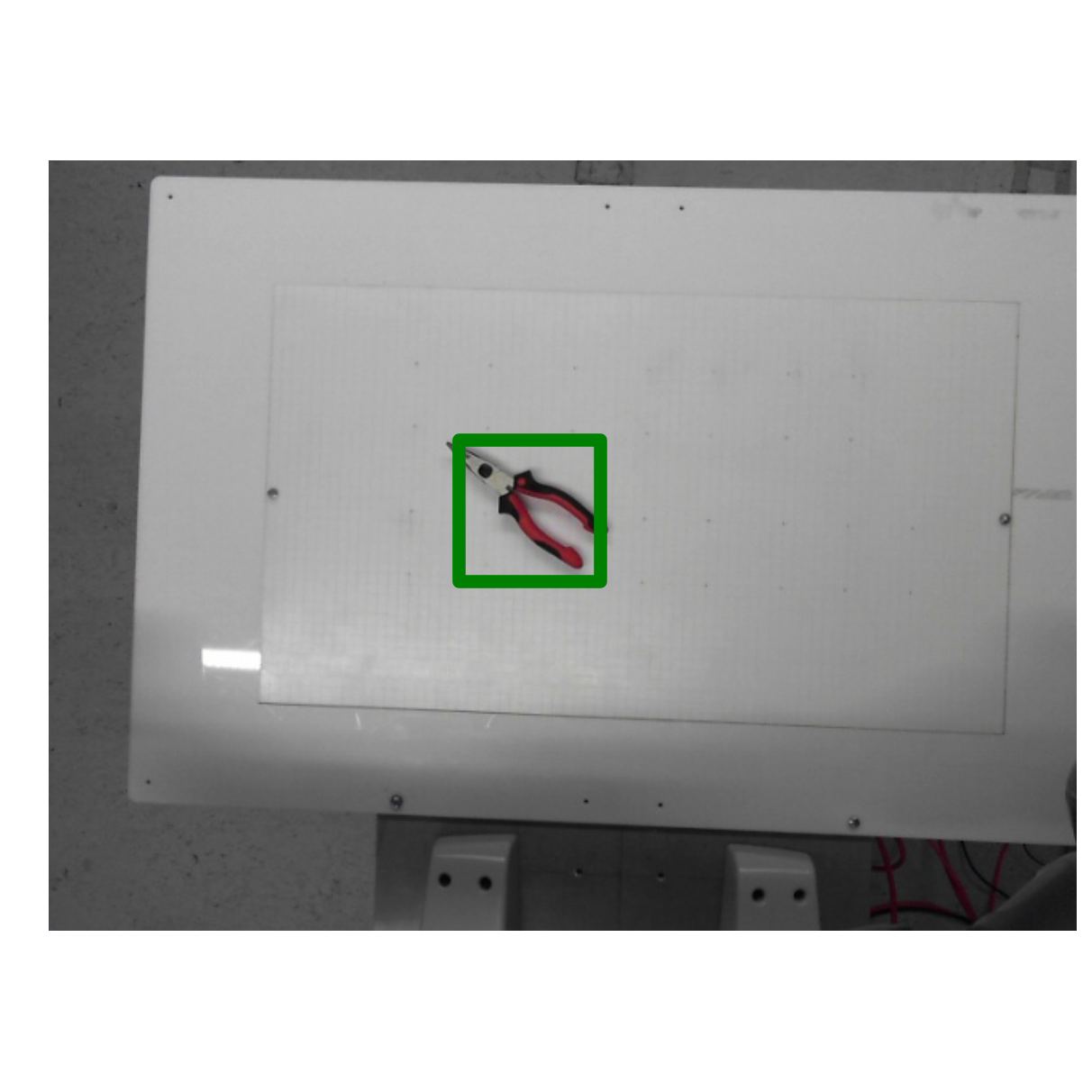}}
                 \quad
                 \subfloat{\includegraphics[width=0.3\columnwidth]{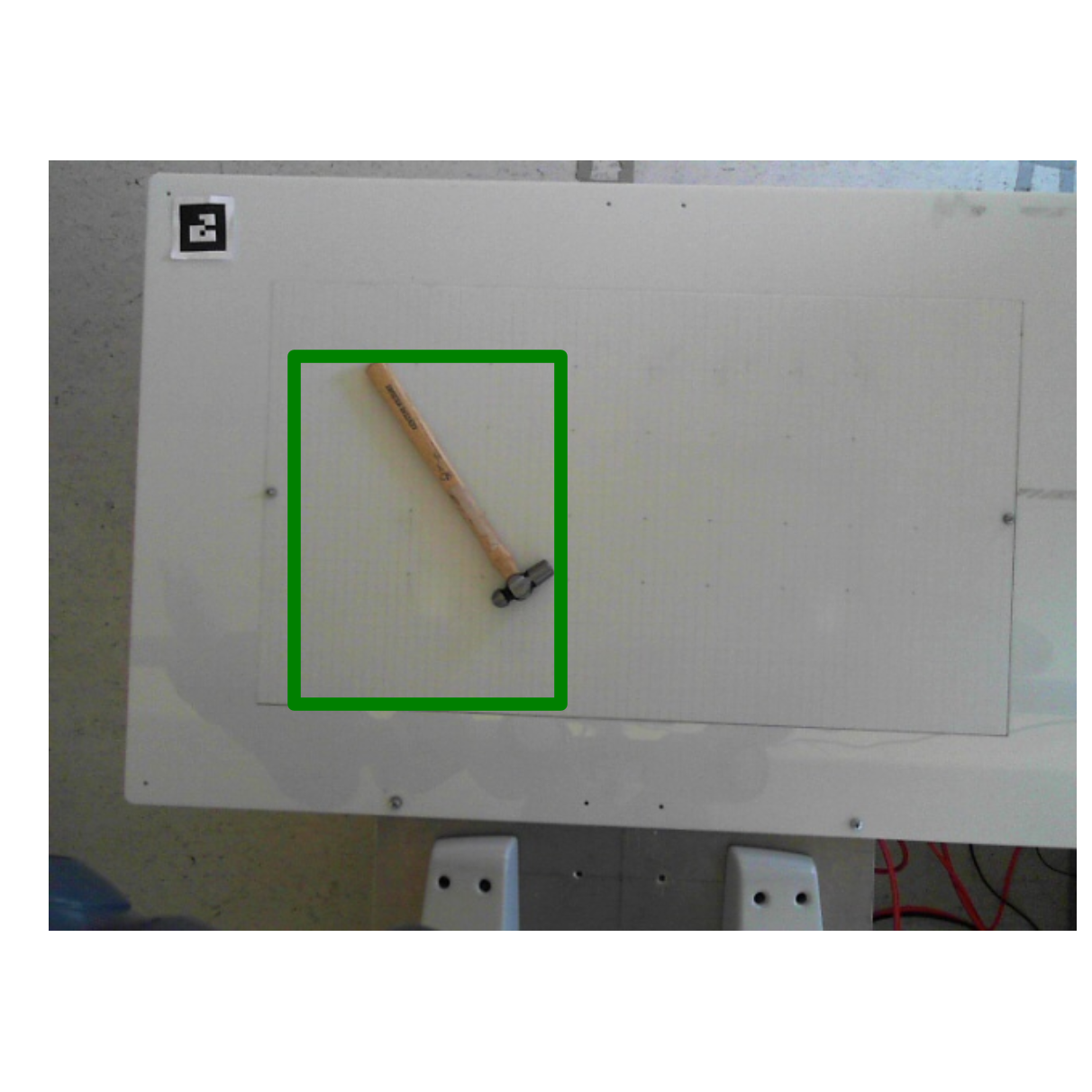}}
                 \quad
                 \subfloat{\includegraphics[width=0.3\columnwidth]{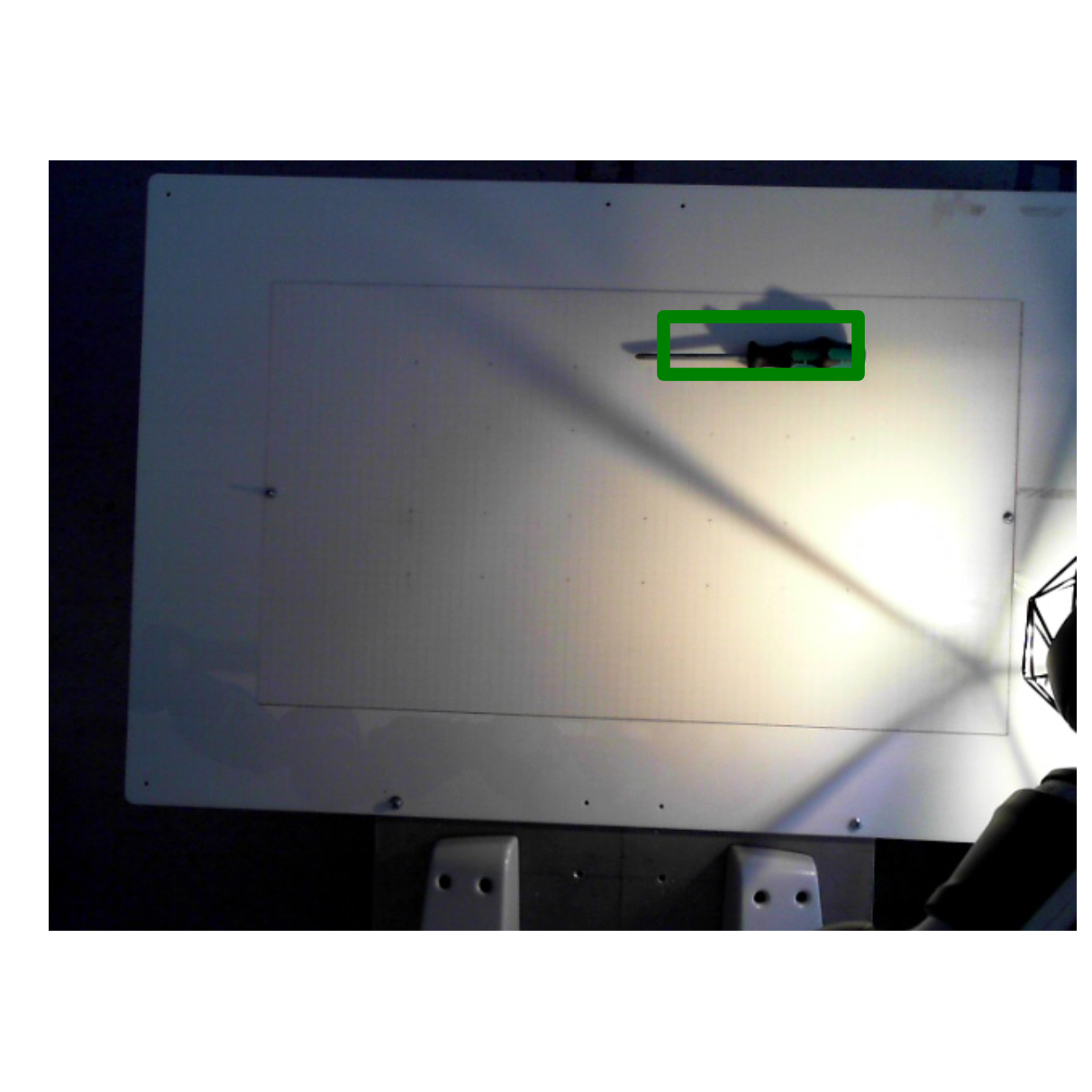}}
         \caption{Example correct localizations from KTH Handtool dataset. The columns from left to right shows artificial, natural and directional illumination. Top row: Results from $Camera_1$. Bottom row: Results from $Camera_2$.}
     \label{fig:kthdata_res}
 \end{figure}

Fig.~\ref{fig:kthdata_res} shows some successful localizations of our system in different illumination conditions and viewpoints.

\section{Conclusion}
\label{sec:conclusion}
In this paper, we have presented an approach for unsupervised object localization in single images by fusing saliency maps and region proposals which are generated by two separate deep neural networks. Our approach performs similar to state-of-the-art methods on a benchmark dataset in terms of average localization accuracy with one-shot processing. We have also evaluated our approach in a challenging dataset with varying environmental conditions and non-dominant objects to get more insights about the performance of the method in real world settings. The datasets used in this work are biased towards localizing one large object in a scene. In case of multiple objects this can cause failures during cue fusion such as the one shown in Table~\ref{tab:objdis_respic} (bottom row). In future work we plan to improve the cue fusion part of our method to be able to distinguish individual object localizations from localizing smaller parts of a bigger object in multi-object localization scenarios. We also plan to investigate how to improve the robustness and precision of the saliency maps w.r.t changing environmental conditions.



\section{Acknowledgements}
The work was funded by the Swedish Foundation for Strategic Research (SSF) through the project FACT and the Centre for Autonomous Systems.

\bibliographystyle{IEEEtran}
\balance
\bibliography{IEEEabrv,bibtex}

\begin{thebibliography}{10}
\providecommand{\url}[1]{#1}
\csname url@rmstyle\endcsname
\providecommand{\newblock}{\relax}
\providecommand{\bibinfo}[2]{#2}
\providecommand\BIBentrySTDinterwordspacing{\spaceskip=0pt\relax}
\providecommand\BIBentryALTinterwordstretchfactor{4}
\providecommand\BIBentryALTinterwordspacing{\spaceskip=\fontdimen2\font plus
\BIBentryALTinterwordstretchfactor\fontdimen3\font minus
  \fontdimen4\font\relax}
\providecommand\BIBforeignlanguage[2]{{%
\expandafter\ifx\csname l@#1\endcsname\relax
\typeout{** WARNING: IEEEtran.bst: No hyphenation pattern has been}%
\typeout{** loaded for the language `#1'. Using the pattern for}%
\typeout{** the default language instead.}%
\else
\language=\csname l@#1\endcsname
\fi
#2}}

\bibitem{lowe1999object}
D.~G. Lowe, ``Object recognition from local scale-invariant features,'' in
  \emph{Computer vision, 1999. The proceedings of the seventh IEEE
  international conference on}, vol.~2.\hskip 1em plus 0.5em minus 0.4em\relax
  Ieee, 1999, pp. 1150--1157.

\bibitem{sullivan1999object}
J.~Sullivan, A.~Blake, M.~Isard, and J.~MacCormick, ``Object localization by
  bayesian correlation,'' in \emph{Computer Vision, 1999. The Proceedings of
  the Seventh IEEE International Conference on}, vol.~2.\hskip 1em plus 0.5em
  minus 0.4em\relax IEEE, 1999, pp. 1068--1075.

\bibitem{imagenet}
J.~Deng, W.~Dong, R.~Socher, L.-J. Li, K.~Li, and L.~Fei-Fei, ``Imagenet: A
  large-scale hierarchical image database,'' in \emph{Proceedings of the IEEE
  conference on computer vision and pattern recognition}.\hskip 1em plus 0.5em
  minus 0.4em\relax IEEE, 2009, pp. 248--255.

\bibitem{microsoft_coco}
T.-Y. Lin, M.~Maire, S.~Belongie, J.~Hays, P.~Perona, D.~Ramanan,
  P.~Doll{\'a}r, and C.~L. Zitnick, ``Microsoft coco: Common objects in
  context,'' in \emph{European conference on computer vision}.\hskip 1em plus
  0.5em minus 0.4em\relax Springer, 2014, pp. 740--755.

\bibitem{Salimans2016}
T.~{Salimans}, I.~{Goodfellow}, W.~{Zaremba}, V.~{Cheung}, A.~{Radford}, and
  X.~{Chen}, ``{Improved Techniques for Training GANs},'' \emph{ArXiv
  e-prints}, June 2016.

\bibitem{cinbis2014multi}
R.~G. Cinbis, J.~Verbeek, and C.~Schmid, ``Multi-fold mil training for weakly
  supervised object localization,'' in \emph{Proceedings of the IEEE conference
  on computer vision and pattern recognition}.\hskip 1em plus 0.5em minus
  0.4em\relax IEEE, 2014, pp. 2409--2416.

\bibitem{deselaers2010}
T.~Deselaers, B.~Alexe, and V.~Ferrari, ``Localizing objects while learning
  their appearance,'' in \emph{European conference on computer vision}.\hskip
  1em plus 0.5em minus 0.4em\relax Springer, 2010, pp. 452--466.

\bibitem{bjorkman2010active}
M.~Bj{\"o}rkman and D.~Kragic, ``Active 3d scene segmentation and detection of
  unknown objects,'' in \emph{Robotics and Automation (ICRA), 2010 IEEE
  International Conference on}.\hskip 1em plus 0.5em minus 0.4em\relax IEEE,
  2010, pp. 3114--3120.

\bibitem{kootstra2011fast}
G.~Kootstra and D.~Kragic, ``Fast and bottom-up object detection, segmentation,
  and evaluation using gestalt principles,'' in \emph{Robotics and Automation
  (ICRA), 2011 IEEE International Conference on}.\hskip 1em plus 0.5em minus
  0.4em\relax IEEE, 2011, pp. 3423--3428.

\bibitem{potapova2014attention}
E.~Potapova, K.~M. Varadarajan, A.~Richtsfeld, M.~Zillich, and M.~Vincze,
  ``Attention-driven object detection and segmentation of cluttered table
  scenes using 2.5 d symmetry,'' in \emph{Robotics and Automation (ICRA), 2014
  IEEE International Conference on}.\hskip 1em plus 0.5em minus 0.4em\relax
  IEEE, 2014, pp. 4946--4952.

\bibitem{ma2015simultaneous}
L.~Ma, M.~Ghafarianzadeh, D.~Coleman, N.~Correll, and G.~Sibley, ``Simultaneous
  localization, mapping, and manipulation for unsupervised object discovery,''
  in \emph{Robotics and Automation (ICRA), 2015 IEEE International Conference
  on}.\hskip 1em plus 0.5em minus 0.4em\relax IEEE, 2015, pp. 1344--1351.

\bibitem{Abbeloos20173DOD}
W.~Abbeloos, E.~A. Cansizoglu, S.~Caccamo, Y.~Taguchi, and Y.~Domae, ``3d
  object discovery and modeling using single rgb-d images containing multiple
  object instances,'' \emph{CoRR}, vol. abs/1710.06231, 2017.

\bibitem{russell2006using}
B.~C. Russell, W.~T. Freeman, A.~A. Efros, J.~Sivic, and A.~Zisserman, ``Using
  multiple segmentations to discover objects and their extent in image
  collections,'' in \emph{Computer Vision and Pattern Recognition, 2006 IEEE
  Computer Society Conference on}, vol.~2.\hskip 1em plus 0.5em minus
  0.4em\relax IEEE, 2006, pp. 1605--1614.

\bibitem{felzenszwalb2004efficient}
P.~F. Felzenszwalb and D.~P. Huttenlocher, ``Efficient graph-based image
  segmentation,'' \emph{International journal of computer vision}, vol.~59,
  no.~2, pp. 167--181, 2004.

\bibitem{Rubinstein2013}
M.~Rubinstein, A.~Joulin, J.~Kopf, and C.~Liu, ``Unsupervised joint object
  discovery and segmentation in internet images,'' in \emph{2013 IEEE
  Conference on Computer Vision and Pattern Recognition}, June 2013, pp.
  1939--1946.

\bibitem{Ghafarianzadeh2014}
M.~Ghafarianzadeh, M.~Blaschko, and G.~Sibley, ``{Unsupervised Spatio-Temporal
  Segmentation with Sparse Spectral Clustering},'' in \emph{{British Machine
  Vision Conference (BMVC)}}, Nottingham, United Kingdom, Sept. 2014.

\bibitem{Cho2015}
M.~{Cho}, S.~{Kwak}, C.~{Schmid}, and J.~{Ponce}, ``{Unsupervised Object
  Discovery and Localization in the Wild: Part-based Matching with Bottom-up
  Region Proposals},'' in \emph{Proceedings of the IEEE conference on computer
  vision and pattern recognition}, Jan. 2015.

\bibitem{Kootstra2010}
G.~Kootstra, N.~Bergström, and D.~Kragic, ``Fast and automatic detection and
  segmentation of unknown objects,'' in \emph{2010 10th IEEE-RAS International
  Conference on Humanoid Robots}, Dec 2010, pp. 442--447.

\bibitem{mishra2009active}
A.~Mishra, Y.~Aloimonos, and C.~L. Fah, ``Active segmentation with fixation,''
  in \emph{Computer Vision, 2009 IEEE 12th International Conference on}.\hskip
  1em plus 0.5em minus 0.4em\relax IEEE, 2009, pp. 468--475.

\bibitem{grundmann2010}
M.~Grundmann, V.~Kwatra, M.~Han, and I.~Essa, ``Efficient hierarchical
  graph-based video segmentation,'' in \emph{2010 IEEE Computer Society
  Conference on Computer Vision and Pattern Recognition}, June 2010, pp.
  2141--2148.

\bibitem{kootstra2010fixation}
G.~Kootstra, N.~Bergstrom, and D.~Kragic, ``Using symmetry to select fixation
  points for segmentation,'' in \emph{Pattern Recognition (ICPR), 2010 20th
  International Conference on}.\hskip 1em plus 0.5em minus 0.4em\relax IEEE,
  2010, pp. 3894--3897.

\bibitem{Horbert2015}
E.~Horbert, G.~M. García, S.~Frintrop, and B.~Leibe, ``Sequence-level object
  candidates based on saliency for generic object recognition on mobile
  systems,'' in \emph{2015 IEEE International Conference on Robotics and
  Automation (ICRA)}, May 2015, pp. 127--134.

\bibitem{garcia2015saliency}
G.~M. Garc{\'\i}a, E.~Potapova, T.~Werner, M.~Zillich, M.~Vincze, and
  S.~Frintrop, ``Saliency-based object discovery on rgb-d data with a
  late-fusion approach,'' in \emph{Robotics and Automation (ICRA), 2015 IEEE
  International Conference on}.\hskip 1em plus 0.5em minus 0.4em\relax IEEE,
  2015, pp. 1866--1873.

\bibitem{Tang2014}
K.~Tang, A.~Joulin, L.-J. Li, and L.~Fei-Fei, ``Co-localization in real-world
  images,'' in \emph{Proceedings of the IEEE conference on computer vision and
  pattern recognition}, 2014, pp. 1464--1471.

\bibitem{joulin2014efficient}
A.~Joulin, K.~Tang, and L.~Fei-Fei, ``Efficient image and video co-localization
  with frank-wolfe algorithm,'' in \emph{European Conference on Computer
  Vision}.\hskip 1em plus 0.5em minus 0.4em\relax Springer, 2014, pp. 253--268.

\bibitem{Borji15}
A.~Borji, M.~M. Cheng, H.~Jiang, and J.~Li, ``Salient object detection: A
  benchmark,'' \emph{IEEE Transactions on Image Processing}, vol.~24, no.~12,
  pp. 5706--5722, Dec 2015.

\bibitem{alexe2012objectness}
B.~Alexe, T.~Deselaers, and V.~Ferrari, ``Measuring the objectness of image
  windows,'' \emph{IEEE transactions on pattern analysis and machine
  intelligence}, vol.~34, no.~11, pp. 2189--2202, 2012.

\bibitem{renNIPS15fasterrcnn}
S.~Ren, K.~He, R.~Girshick, and J.~Sun, ``Faster {R-CNN}: Towards real-time
  object detection with region proposal networks,'' in \emph{Advances in Neural
  Information Processing Systems ({NIPS})}, 2015.

\bibitem{Yu2017}
F.~Yu, V.~Koltun, and T.~Funkhouser, ``Dilated residual networks,'' in
  \emph{Computer Vision and Pattern Recognition (CVPR)}, 2017.

\bibitem{Yu2015}
F.~{Yu}, A.~{Seff}, Y.~{Zhang}, S.~{Song}, T.~{Funkhouser}, and J.~{Xiao},
  ``{LSUN: Construction of a Large-scale Image Dataset using Deep Learning with
  Humans in the Loop},'' \emph{ArXiv e-prints}, June 2015.

\bibitem{dai16rfcn}
D.~Jifeng, L.~Yi, H.~Kaiming, and J.~Sun, ``{R-FCN}: Object detection via
  region-based fully convolutional networks,'' \emph{arXiv preprint
  arXiv:1605.06409}, 2016.

\bibitem{kth_handtool}
\BIBentryALTinterwordspacing
``Kth handtool dataset,'' (Date last accessed 11-April-2018). [Online].
  Available: \url{https://www.nada.kth.se/cas/data/handtool/}
\BIBentrySTDinterwordspacing

\end{thebibliography}

\end{document}